\documentclass[11pt,a4paper]{article}

\usepackage[margin=1in]{geometry}
\usepackage{amsmath,amssymb,amsthm}
\usepackage{graphicx}
\usepackage{booktabs}
\usepackage{multirow}
\usepackage{array}
\usepackage{xcolor}
\usepackage{hyperref}
\usepackage{pgfplots}
\pgfplotsset{compat=1.18}
\usepackage{tikz}
\usetikzlibrary{shapes.geometric, arrows.meta, positioning, fit, backgrounds}
\usepackage{algorithm}
\usepackage{algpseudocode}
\usepackage{caption}
\usepackage{subcaption}
\usepackage[numbers]{natbib}
\usepackage{microtype}
\usepackage{enumitem}
\usepackage{seqsplit}
\usepackage{longtable}

\newcommand{\xmark}{\ensuremath{\times}}

\newtheorem{definition}{Definition}
\newtheorem{hypothesis}{Hypothesis}
\newtheorem{principle}{Principle}

\definecolor{verA}{RGB}{50,102,173}
\definecolor{verB}{RGB}{29,158,117}
\definecolor{verC}{RGB}{216,90,48}
\definecolor{verD}{RGB}{186,117,23}
\definecolor{lightgray}{RGB}{240,240,240}

\title{\textbf{Self-Describing Structured Data with Dual-Layer Guidance:\\
A Lightweight Alternative to RAG for Precision Retrieval\\
in Large-Scale LLM Knowledge Navigation}}

\author{Liu Hung Ming\thanks{PARRAWA AI} \\ {cyril.liu@gmail.com}}

\date{March 2026}

\begin{document}
\maketitle

\begin{abstract}
Large Language Models (LLMs) exhibit a well-documented positional bias when
processing long input contexts: information placed in the middle of a large
context window receives substantially less attention than information at the
beginning or end, a phenomenon termed the \emph{Lost-in-the-Middle} effect
\citep{liu2024lost}. This limitation poses a significant challenge for
knowledge-retrieval applications that embed large structured knowledge bases
directly in the LLM context. Retrieval-Augmented Generation (RAG) addresses
scalability by retrieving only relevant fragments, but introduces substantial
infrastructure overhead and is ill-suited to structured knowledge libraries
whose semantic boundaries are human-defined rather than statistically learned.

We propose \textbf{Self-Describing Structured Retrieval} (SDSR), a
lightweight framework in which structured data files embed their own
human-authored navigational metadata (\texttt{\_summary} block with a
\texttt{category\_index} and per-category \texttt{routing\_hint} fields)
at the file's primacy position, thereby exploiting the LLM's primacy bias
rather than fighting it. We further identify that in-file guidance alone is
insufficient at large scale, and propose a complementary
\textbf{Dual-Layer Guidance} strategy that combines in-file metadata with
explicit abstract-level routing rules in the system prompt.

We validate this framework through a controlled four-round benchmark
experiment using a 190-skill knowledge library that we systematically expand
from 36 to 119 categories by injecting semantically adversarial distractor
categories. Four experimental conditions are tested: (A)~no guidance,
(B)~in-file summary only, (C)~prompt hint only, and (D)~both combined.
Version~D achieves \textbf{100\% primary routing accuracy} (20/20) at the
119-category scale, compared to 65\% for the no-guidance baseline.
Our analysis reveals a fundamental asymmetry between primary routing
(solvable by explicit rules) and secondary cross-category routing
(requiring implicit architectural intent encoded in data structure), with
implications for LLM knowledge base design. We further extend SDSR to a general architecture for high-precision
semantic retrieval over large semi-structured corpora, demonstrating how
a one-time structuring pass with cross-reference encoding enables SDSR
to operate on domains with recoverable document structure---such as legal
judgments---without vector databases or embedding infrastructure.
\end{abstract}


\newpage

\section{Introduction}
\label{sec:intro}

The ability of Large Language Models to process long input contexts has
expanded dramatically, with contemporary models supporting context windows ranging from
128K to over one million tokens. However, empirical studies consistently
demonstrate that effective utilisation degrades well before the stated
limit: models do not uniformly attend to their full input, and
performance on retrieval-dependent tasks can drop by over 30\% when
relevant information shifts from boundary positions to the middle of the
context \citep{liu2024lost}.  
Yet the practical utility of these extended windows is limited by a
structural property of transformer attention: models attend disproportionately
to tokens near the beginning and end of their input, with information
in the middle receiving systematically weaker signal. \citet{liu2024lost}
formalised this as the \emph{Lost-in-the-Middle} effect, demonstrating
a U-shaped performance curve across multiple LLMs and tasks.

This positional bias creates a practical dilemma for knowledge-intensive
applications. Consider a practitioner who maintains a structured library of
domain-specific skill specifications---a knowledge base that has grown to
hundreds of categorised entries, each with precise semantic boundaries defined
through expert curation. Embedding the entire library in a single LLM
context should, in theory, enable the model to navigate the full space of
available knowledge. In practice, the model's attention degrades as library
size grows, and carefully designed distinctions between adjacent categories
become invisible to it.

The dominant engineering response to this problem is
Retrieval-Augmented Generation (RAG) \citep{lewis2020rag}: rather than
providing the full knowledge base, a retrieval system selects semantically
relevant fragments and injects only those into the context. RAG is highly
effective for unstructured text corpora but carries substantial costs:
it requires chunking strategies that may sever semantic units,
embedding models that encode surface similarity rather than human-defined
boundaries, and vector databases that must be maintained as the knowledge
base evolves. For knowledge bases whose structure is explicitly human-designed,
these costs are unnecessary---the semantic boundaries are already known.

In this paper, we ask: \emph{can a structured knowledge file be designed to
guide its own reading by an LLM, exploiting rather than fighting the
positional bias?}

We answer this question through a series of controlled experiments motivated
by an empirical observation: when navigational instructions (a summary index
with per-category routing hints) are embedded at the beginning of a structured
JSON knowledge file, the LLM's routing accuracy improves---not because the
instructions are in the system prompt, but because they occupy the high-primacy
position within the data object being processed. This observation suggests
that the LLM treats the data file as a distinct cognitive object from the
system prompt, and that instructions embedded within that object benefit
from a different attentional regime.

\paragraph{Contributions.} This paper makes the following contributions:

\begin{enumerate}[leftmargin=*, label=\arabic*.]
  \item We introduce the \textbf{Self-Describing Structured Retrieval (SDSR)}
        framework, in which structured data files embed human-authored
        navigational metadata at their primacy position to guide LLM
        navigation without external retrieval infrastructure.

  \item We conduct a systematic \textbf{four-round benchmark experiment}
        with controlled category-count scaling (36 $\to$ 60 $\to$ 119
        categories) and adversarial distractor injection, comparing four
        guidance conditions across all scales.

  \item We identify and formalise the \textbf{Dual-Layer Guidance} principle:
        in-file primacy metadata handles structural navigation while prompt-level
        abstract routing rules handle abstraction-level disambiguation---the
        two layers are complementary and non-redundant.

  \item We characterise the \textbf{primary/secondary routing asymmetry}:
        primary category selection is solvable by explicit guidance rules,
        while secondary cross-category pairing requires implicit architectural
        intent that must be explicitly encoded in data structure fields
        (e.g., a \texttt{complement} field).

  \item We extend SDSR to a general architecture for \textbf{high-precision
        retrieval over large structured corpora}, including a two-tier
        Python-controlled reading pipeline applicable to domains such as
        legal document retrieval.
\end{enumerate}

\section{Related Work}
\label{sec:related}

\subsection{The Lost-in-the-Middle Effect}

\citet{liu2024lost} conducted controlled experiments measuring LLM performance
on multi-document question answering and key-value retrieval as a function of
the position of relevant information within the input context. They found a
characteristic U-shaped performance curve: accuracy is highest when relevant
information appears at the very beginning (primacy bias) or end (recency bias)
of the context, and degrades significantly for information in the middle.
This effect persists across both open-source models (MPT-30B, LongChat-13B)
and closed models (GPT-3.5-Turbo, Claude-1.3), and does not disappear
for models explicitly designed for long-context processing.

The authors connect this finding to the \emph{serial-position effect} from
cognitive psychology \citep{ebbinghaus1913}, observing that the U-shaped curve
mirrors the recall patterns observed in human free-association memory tasks.
Importantly for our work, they also find that \emph{query-aware
contextualization}---placing the query both before and after the document
list---improves performance on synthetic retrieval tasks, providing preliminary
evidence that structural positioning of navigational content can modulate
attention allocation.

Subsequent work has extended this analysis to longer contexts and more
complex tasks \citep{he2024position,fu2024data}, consistently confirming
that even models with context windows of 128K tokens do not uniformly
attend to their full input. From a practical standpoint,
\citet{he2024position} show that reranking retrieved documents to
place the most relevant content at the beginning or end of the
context---a strategy they term \emph{position engineering}---can yield
substantial accuracy improvements at zero additional inference cost.

Two recent studies provide deeper theoretical grounding for the
positional biases that SDSR exploits. \citet{wu2025emergence} present a
graph-theoretic framework showing that primacy bias arises from two
architectural sources: causal attention masking, which causes tokens in
deeper layers to attend to increasingly contextualised representations
of earlier positions, and the interaction between causal masking and
relative positional encodings (notably RoPE), which produces a
trade-off between long-term decay and the cumulative importance of
early sequence positions. Their Theorem~4.1 establishes that primacy
bias is a mathematical consequence of causal masking in multi-layer
attention, not merely an empirical regularity---providing an
architectural guarantee for SDSR's strategy of placing navigational
metadata at the file's primacy position.

Complementing this architectural analysis, \citet{salvatore2025lost}
argue that the U-shaped performance curve is not a defect but an
\emph{emergent adaptation} to competing information-retrieval demands
in pre-training data: tasks requiring uniform recall across the full
input (long-term memory demand) interact with causal masking to produce
primacy effects, while tasks prioritising recent information produce
recency effects. Crucially, they demonstrate that the primacy effect
emerges in autoregressive architectures (GPT-2, Llama, RNNs) but not
in bidirectional encoder-decoders (T5), suggesting that SDSR's
primacy-exploitation strategy is robust for the current generation of
decoder-only LLMs but may require reassessment if future dominant
architectures adopt bidirectional attention.

\subsection{Retrieval-Augmented Generation}

Retrieval-Augmented Generation was introduced by \citet{lewis2020rag} as a
framework for grounding LLM generation in external knowledge bases without
retraining. The canonical architecture retrieves the top-$k$ document
chunks from a dense vector index using a learned query encoder, then
concatenates these chunks with the user query as context for a
generative model.

RAG has since been extended in multiple directions. \citet{yu2024rankrag}
unify context ranking with generation by fine-tuning a single model to
simultaneously rank retrieved passages and generate answers.
Graph-augmented RAG (GraphRAG) \citep{edge2024graphrag,guo2024survey}
addresses the loss of inter-document relationships inherent in flat
chunk retrieval by constructing knowledge graphs over the document corpus
and retrieving over graph neighbourhoods rather than independent chunks.
Hierarchical RAG \citep{chen2024benchmarking} preserves document structure
through multi-level retrieval, partially addressing the concern that
fixed-size chunking severs semantic units.

Despite these advances, RAG architectures share a common limitation for
our target domain: they are optimised for \emph{statistical} semantic
similarity, encoding the surface-level co-occurrence patterns of their
training corpus. When a knowledge base has explicit, human-defined semantic
boundaries that may not correspond to surface co-occurrence---as is the case
for expert-curated skill libraries, legal taxonomies, or medical ontologies
---vector similarity retrieval may systematically prefer the wrong fragment.
\citet{packowski2024optimizing} report from their experience building
enterprise-scale RAG solutions that achieving stable and accurate results
for complex queries requires substantial content engineering effort,
including iterative optimisation of the knowledge base content itself
rather than relying solely on retrieval-side tuning.

\subsection{Context Engineering and Prompt Positioning}

A growing body of work studies how the structural properties of the
input context affect LLM behaviour beyond simple retrieval accuracy.
Beyond reranking, \citet{he2024position} further show that directly
manipulating positional indices---without altering the prompt text
itself---can improve RAG and in-context learning performance
substantially, demonstrating that positional information is a
first-class design variable. \citet{guo2024serial} demonstrate that
LLMs exhibit serial position effects---primacy and recency
biases---analogous to those documented in human free-recall
experiments, arguing that prompt design should account for these
biases rather than assuming uniform attention.

Meta-prompting \citep{suzgun2024metaprompting} explores using one LLM
to generate structured prompts for another, demonstrating that explicit
structural framing can improve downstream task accuracy by 15--17\%.
The Prompt Report \citep{schulhoff2024prompt} surveys over 50 distinct
prompting techniques, noting that techniques which provide explicit
structural scaffolding consistently outperform those relying on the model's
implicit knowledge organisation.

\subsection{Knowledge Base Design for LLM Consumption}

Relatively little work addresses the question of how structured knowledge
bases should be \emph{designed} to be navigated effectively by LLMs.
\citet{packowski2024optimizing} study knowledge base content design for
enterprise RAG, focusing on the format of text conversions from PDF and HTML,
but do not address inline navigational metadata.

To the best of our knowledge, no prior work has proposed embedding
human-authored navigational metadata directly within structured data files
as a mechanism for exploiting positional bias during LLM knowledge
navigation. The present work fills this gap.

\section{Problem Formulation}
\label{sec:problem}

\subsection{Structured Knowledge Library Navigation}

\begin{definition}[Structured Knowledge Library]
A \emph{structured knowledge library} $\mathcal{L}$ is a collection of
$N$ categories $\{C_1, C_2, \ldots, C_N\}$, where each category $C_i$
contains a name $n_i$, a description $d_i$, and a set of skill entries
$S_i = \{s_{i,1}, \ldots, s_{i,k_i}\}$. Each skill entry $s_{i,j}$ has
a name and optionally a description.
\end{definition}

\begin{definition}[Routing Task]
Given a task description $q$ and a library $\mathcal{L}$, the
\emph{primary routing task} $R_1(q, \mathcal{L})$ requires selecting the
category $C^* \in \mathcal{L}$ that best addresses $q$. The
\emph{secondary routing task} $R_2(q, \mathcal{L}, C^*)$ requires selecting
a complementary category $C^{**} \neq C^*$ that addresses a distinct
architectural dimension of $q$.
\end{definition}

\begin{definition}[Distractor Category]
A \emph{distractor category} $C^D$ is a category whose name and description
are semantically proximate to a target category $C^*$ in surface keyword
space but occupy a different abstraction level (mechanism vs.\ governance,
component vs.\ pipeline). Formally, $\text{sim}_\text{surface}(C^D, C^*) >
\theta$ while $\text{level}(C^D) \neq \text{level}(C^*)$.
\end{definition}

\subsection{Research Hypotheses}

We investigate three nested hypotheses about the effectiveness of
different guidance strategies:

\begin{hypothesis}[Self-Describing Data]\label{hyp:infile}
Embedding a human-authored \texttt{\_summary} block with
\texttt{routing\_hint} fields at the beginning of a structured data file
improves LLM primary routing accuracy compared to providing no guidance,
particularly as library size increases.
\end{hypothesis}

\begin{hypothesis}[Prompt Guidance]\label{hyp:prompt}
Providing explicit routing rules in the system prompt---including explicit
category naming and abstract-level selection rules---improves primary routing
accuracy compared to providing no guidance.
\end{hypothesis}

\begin{hypothesis}[Dual-Layer Synergy]\label{hyp:dual}
Combining in-file metadata (Hypothesis~\ref{hyp:infile}) with prompt-level
routing rules (Hypothesis~\ref{hyp:prompt}) produces a synergistic improvement
in primary routing accuracy that exceeds either strategy alone, because the
two layers serve complementary functions: in-file metadata provides structural
navigation at the data's primacy position, while prompt rules provide
abstraction-level disambiguation.
\end{hypothesis}

\subsection{Scoring Protocol}

For each task $q_t$ ($t = 1, \ldots, 20$), an LLM response is scored as:

\begin{equation}
\text{score}(t) =
\begin{cases}
1.0 & \text{if primary category is correct} \\
1.5 & \text{if both primary and secondary categories are correct} \\
0.0 & \text{otherwise}
\end{cases}
\end{equation}

The maximum achievable score is $20.0 + 8.5 \times 0.5 = 28.5$ points,
where 8.5 reflects the number of questions with a designated secondary
category in the answer key.

Primary accuracy is reported as $\text{PA} = \frac{|\{t : C^*_t \text{ correct}\}|}{20}$
and secondary hit rate as $\text{SHR} = \frac{|\{t : C^{**}_t \text{ correct}\}|}{|\{t : C^{**}_t \text{ defined}\}|}$.

\section{Methodology}
\label{sec:methodology}

\subsection{Knowledge Library: High-Impact Skills Library}

Our experiments use the \emph{High-Impact Skills Library} (HISL), a
practitioner-developed prompt-engineering knowledge base comprising
36 human-curated categories and 190 skill entries.
Each category represents a distinct capability domain (e.g.,
\texttt{Cognitive\_Architecture\_\&\_Routing},
\texttt{Axiomatic\_Logic\_\&\_Audit\_Systems},
\texttt{Academic\_Research\_Synthesis\_Pipeline}),
with an expert-authored \texttt{category\_description} that defines
the category's scope and distinguishes it from adjacent categories.
Each skill entry contains a \texttt{skill\_name} and a one-to-two sentence
\texttt{description} specifying its function within the category.

The library is serialised as a single JSON file.
The \emph{Version~B} (in-file summary) variant prepends a
\texttt{\_summary} block before the main \texttt{High\_Impact\_Skills\_Library}
object, containing: (i)~a \texttt{category\_index} listing all categories
with their skill counts and a 100-character \texttt{routing\_hint} excerpted
from each category description; (ii)~\texttt{\_llm\_instructions} specifying
how to use the index for two-stage navigation; and (iii)~\texttt{routing\_roles}
mapping meta-functions (cognitive anchor, universal fallback, domain-specific)
to responsible categories.

\subsection{Experimental Conditions}

We define four experimental conditions that vary the type and placement of
navigational guidance:

\begin{description}
  \item[\textbf{Version A} (No Guidance).] The bare JSON library is provided
    without any \texttt{\_summary} block. The system prompt contains only a
    minimal professional framing with no structural instructions.
    This condition establishes the no-guidance baseline.

  \item[\textbf{Version B} (In-File Summary Only).] The JSON library includes
    the full \texttt{\_summary} block with \texttt{category\_index} and
    \texttt{routing\_hint} fields at the file's primacy position.
    The system prompt is identical to Version~A.

  \item[\textbf{Version C} (Prompt Hint Only).] The bare JSON library
    (identical to Version~A) is provided. The system prompt contains
    explicit structural guidance: a list of key categories with
    their roles, an abstraction-level priority rule
    (\emph{``prefer broader pipeline/governance categories over
    narrower mechanism categories when both seem relevant''}),
    and instructions to use the \texttt{category\_description}
    fields to confirm relevance before selecting.

  \item[\textbf{Version D} (Dual-Layer Guidance).] Combines the in-file
    \texttt{\_summary} block of Version~B with the prompt-level routing
    rules of Version~C. The system prompt additionally names the seven
    highest-priority categories that are most vulnerable to distractor
    confusion (Table~\ref{tab:high_priority_cats}).
\end{description}

\begin{table}[t]
\centering
\caption{High-priority categories explicitly named in the Version~D system
         prompt, with their commonly confused distractor categories.}
\label{tab:high_priority_cats}
\small
\begin{tabular}{lll}
\toprule
\textbf{Target Category} & \textbf{Confused With} & \textbf{Confusion Type} \\
\midrule
Axiomatic\_Logic\_\&\_Audit\_Systems      & Recursive\_Self\_Audit\_Engine       & hub vs.\ satellite \\
Distributed\_Cognition\_\&\_Context\_Orch.& Agent\_Handoff\_Protocol\_Design     & governance vs.\ mechanics \\
Adversarial\_Systems\_Thinking           & Competitive\_Intelligence\_Synthesis & framework vs.\ intelligence \\
Academic\_Research\_Synthesis\_Pipeline  & Code\_To\_Methodology\_Translator    & pipeline vs.\ component \\
Revenue\_Generation\_\&\_Commercial\_Logic& Conversion\_Funnel\_Architecture     & system vs.\ mechanism \\
Cross\_Cultural\_Localization\_Intelligence& Cultural\_Signal\_Detection         & system vs.\ detection \\
Interactive\_Narrative\_\&\_Fiction\_Engine& Branch\_Narrative\_Architect        & engine vs.\ component \\
\bottomrule
\end{tabular}
\end{table}

\subsection{Test Question Design}

We design 20 task-description questions $\{q_1, \ldots, q_{20}\}$ to test
routing accuracy across the full range of HISL categories.
A central design principle is \emph{keyword avoidance}: no question
contains any substring of its target category's name, forcing the model to
route by semantic understanding rather than keyword matching.

Questions are stratified by expected discrimination power:

\begin{itemize}
  \item \textbf{High-discrimination questions} (Q09, Q11, Q12, Q16, Q20):
    Target categories are either rare (1--2 skills), positioned late in the
    file, or have names that are non-obvious relative to the task description.
    These are expected to show the largest A/B/C/D divergence.

  \item \textbf{Low-discrimination questions} (Q01, Q05, Q07):
    Target categories have highly intuitive names. All conditions are expected
    to answer correctly; these establish the baseline and verify that guidance
    does not degrade already-correct responses.
\end{itemize}

Table~\ref{tab:question_sample} shows representative questions.

\begin{table}[t]
\centering
\caption{Representative test questions (keyword avoidance enforced).
         Full question set in Appendix~\ref{app:questions}.}
\label{tab:question_sample}
\small
\begin{tabular}{p{0.05\linewidth} p{0.55\linewidth} p{0.30\linewidth}}
\toprule
\textbf{Q\#} & \textbf{Task Description (excerpt)} & \textbf{Primary Target Category} \\
\midrule
Q02 & ``\ldots rigorously check its own output for logical contradictions and audit its own reasoning process step by step.'' & Axiomatic\_Logic\_\&\_Audit\_Systems \\
Q12 & ``\ldots multi-agent AI workflow where each agent handles a different stage; context must be reliably passed between agents without loss.'' & Distributed\_Cognition\_\&\_Context\_Orch. \\
Q14 & ``\ldots automatically convert experimental code results into methodology and results sections formatted for academic publication.'' & Academic\_Research\_Synthesis\_Pipeline \\
Q20 & ``\ldots sound design AI that, given a visual scene description, can derive the corresponding sound layers and acoustic landscape structure.'' & Sensory\_Audio\_Design \\
\bottomrule
\end{tabular}
\end{table}

\subsection{Library Scale Expansion and Distractor Injection}

To test hypotheses across varying levels of navigational challenge,
we expand the library across three rounds:

\begin{description}
  \item[\textbf{Round 1}:] Original HISL, 36 categories, 190 skills.
  \item[\textbf{Round 2}:] 24 distractor categories injected, total 60 categories, 262 skills.
  \item[\textbf{Round 3}:] 60 additional distractor categories injected, total 119 categories, 380 skills.
\end{description}

Distractor categories are designed in two interference tiers:

\begin{itemize}
  \item \textbf{High-interference distractors} (30 categories in Round~3):
    Names are semantically proximate to real target categories
    (e.g., \texttt{Agent\_Handoff\_Protocol\_Design} near
    \texttt{Distributed\_Cognition\_\&\_Context\_Orchestration}).
    Each contains 2--3 skill entries with one-sentence descriptions.

  \item \textbf{Low-interference distractors} (30 categories in Round~3):
    Names use domain analogies unrelated to any test question target
    (e.g., \texttt{Mycological\_Network\_Design},
    \texttt{Trophic\_Cascade\_Analyzer}), providing pure
    volumetric pressure without semantic interference.
\end{itemize}

Distractor categories are interleaved uniformly among real categories
(one distractor inserted after every real category) to prevent positional
clustering effects. The answer key remains anchored to the original 36 real
categories throughout all rounds.

\subsection{Evaluation Protocol}

For each condition $v \in \{A, B, C, D\}$ and round $r \in \{1, 2, 3\}$,
we submit all 20 questions to the LLM in a single fresh conversation,
preceded by the condition-appropriate system prompt and file upload.
The model is instructed to respond in the format
\texttt{Q\#: category\_name | skill\_name} for up to two selections per
question, withholding any chain-of-thought reasoning until all 20 answers
are listed. The answer key is provided in a second turn within the same
conversation, and the model self-scores its responses into a structured
table. All experiments use Claude Opus~4.6 (\texttt{claude-opus-4-6}),
the highest-capability model in the Claude~4.6 family, accessed
via the claude.ai web interface with default inference parameters.
The use of the strongest available model is a deliberate methodological
choice: any observed failures in secondary routing cannot be attributed
to insufficient model capacity, and instead reflect a fundamental
limitation in the availability of the required knowledge within the
data structure.

Condition~D is evaluated only in Round~3 (the most challenging scale),
motivated by the unexpected finding from Round~3 that Version~B $\approx$
Version~A---establishing that in-file guidance alone is insufficient and
a combined strategy is warranted.

\section{Experiments and Results}
\label{sec:experiments}

\subsection{Round 1: Baseline at Original Scale (36 Categories)}

\paragraph{Results.}
All three conditions (A, B, C) achieve perfect primary accuracy
(PA = 20/20, 100\%) in Round~1, with the following secondary hit rates:
A: 2/17 (11.8\%), B: 3/17 (17.6\%), C: 2/17 (11.8\%).
Total scores: A = 21.0/28.5, B = 21.5/28.5, C = 21.0/28.5.

\paragraph{Analysis.}
The perfect primary accuracy across all conditions indicates that at 36
categories, the navigational challenge is insufficient to expose attentional
limits. The HISL's original category names are sufficiently distinctive that
keyword-level matching resolves all 20 primary routing decisions correctly
regardless of guidance condition. The marginal advantage of Version~B
(+0.5 pts from one additional secondary hit in Q18) is not statistically
meaningful given the single-run experimental design.

This round establishes two important baselines: (i)~the answer key is
navigable in principle, ruling out task design flaws; and (ii)~secondary
routing is systematically more difficult than primary routing---a
pattern that persists across all rounds.

\subsection{Round 2: Medium Scale with Adversarial Distractors (60 Categories)}

\paragraph{Results.}
Primary accuracy drops for all conditions relative to Round~1, with the
no-guidance baseline (A) falling to 15/20 (75\%). Both B and C maintain
higher accuracy (B: 16/20 = 80\%, C: 17/20 = 85\%). Secondary hit rates
remain low across all conditions (A: 2/17, B: 3/17, C: 2/17).
Total scores: A = 15.0/20.0 (primary only), B = 16.5/28.5, C = 16.5/28.5.

\paragraph{Analysis.}
The first meaningful divergence between A and the guided conditions
emerges in Round~2, supporting Hypotheses~\ref{hyp:infile}
and~\ref{hyp:prompt}. However, B~$\approx$~C (tied at 16.5/28.5),
meaning that in-file summary and prompt-level guidance are interchangeable
in their effectiveness at this scale.

The five complete misses (Q12, Q13, Q15, Q16, and the reduced Q19)
share a common failure mode: the distractor category name is a
\emph{surface-literal match} to the task description, while the
correct answer requires recognising a higher abstraction level.
For example, Q12 asks about ``reliable context passing between agents''
and the model consistently selects \texttt{Agent\_Handoff\_Protocol\_Design}
(mechanism level) rather than
\texttt{Distributed\_Cognition\_\&\_Context\_Orchestration}
(governance level).

\subsection{Round 3: Large Scale (119 Categories)}

\paragraph{Results.}
At 119 categories, the gap between conditions widens substantially.
Version~A falls to 13/20 (65\%), Version~B also falls to 13/20 (65\%),
while Version~C holds at 16/20 (80\%).
The unexpected result is that B~$=$~A: the in-file summary provides
\emph{no benefit} over no guidance at this scale.

Total scores: A = 13.0, B = 13.0, C = 16.0 (primary hits only,
secondary = 0 for all conditions).

\paragraph{Analysis.}
The equivalence of B and A at 119 categories reveals the scalability
limit of the in-file summary approach. With 119 routing hints packed
into the \texttt{category\_index}, each hint at 100 characters, the
index itself becomes a dense block of text---subject to the same
lost-in-the-middle degradation it was designed to mitigate.
The routing hints are present and syntactically accessible, but the
LLM's ability to use them for disambiguation degrades as the hint
count grows.

Version~C's advantage over B reveals an important asymmetry:
\emph{natural-language rules in the system prompt} maintain effectiveness
at larger scales because they do not grow in length with the library size.
The Version~C prompt names seven specific high-risk categories and states
one abstract selection rule; this fixed-length, high-salience instruction
is processed in the system prompt's primacy position, not within the
data object.

Table~\ref{tab:full_results} summarises primary accuracy across all
rounds and conditions.

\begin{table}[t]
\centering
\caption{Primary routing accuracy (out of 20) across all experimental
         conditions and rounds. Version~D was evaluated in Round~3 only.}
\label{tab:full_results}
\begin{tabular}{lcccc}
\toprule
\textbf{Round (Categories)} & \textbf{A} & \textbf{B} & \textbf{C} & \textbf{D} \\
\midrule
Round 1 (36) & 20 & 20 & 20 & --- \\
Round 2 (60) & 15 & 16 & 17 & --- \\
Round 3 (119) & 13 & 13 & 16 & \textbf{20} \\
\bottomrule
\end{tabular}
\end{table}

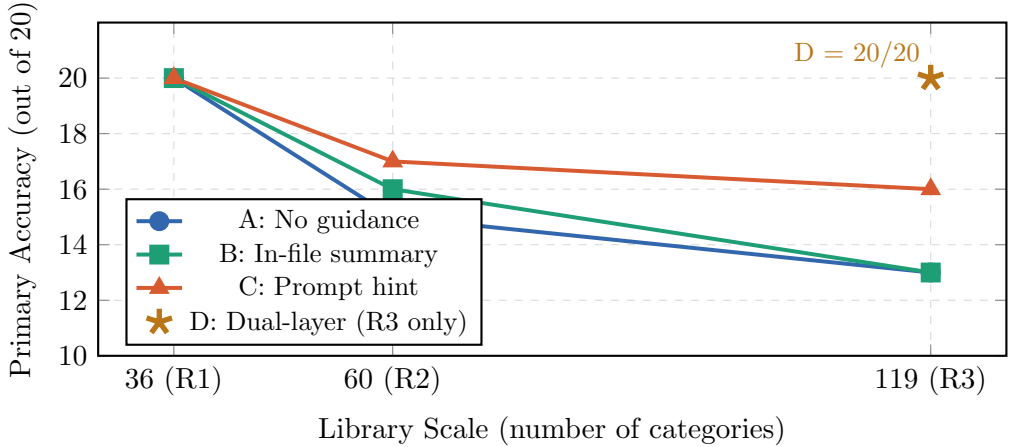
\begin{figure}[t]
\centering
\begin{tikzpicture}
\begin{axis}[
  width=0.85\linewidth,
  height=6cm,
  xlabel={Library Scale (number of categories)},
  ylabel={Primary Accuracy (out of 20)},
  xtick={36,60,119},
  xticklabels={36 (R1), 60 (R2), 119 (R3)},
  ytick={10,12,14,16,18,20},
  ymin=10, ymax=22,
  legend pos=south west,
  legend style={font=\small},
  grid=major,
  grid style={dashed, gray!30},
  thick,
]
\addplot[color=verA, mark=*, mark size=3pt, line width=1.5pt]
  coordinates {(36,20) (60,15) (119,13)};
\addlegendentry{A: No guidance}

\addplot[color=verB, mark=square*, mark size=3pt, line width=1.5pt]
  coordinates {(36,20) (60,16) (119,13)};
\addlegendentry{B: In-file summary}

\addplot[color=verC, mark=triangle*, mark size=3pt, line width=1.5pt]
  coordinates {(36,20) (60,17) (119,16)};
\addlegendentry{C: Prompt hint}

\addplot[color=verD, mark=star, mark size=5pt, only marks, line width=2pt]
  coordinates {(119,20)};
\addlegendentry{D: Dual-layer (R3 only)}

\node[above left, font=\small, color=verD] at (axis cs:119,20) {D = 20/20};

\end{axis}
\end{tikzpicture}
\caption{Primary routing accuracy as a function of library scale for all
         four experimental conditions. Version~D achieves perfect accuracy
         at 119 categories. Versions~A and~B converge at Round~3,
         demonstrating the scalability limit of in-file routing hints alone.}
\label{fig:trend}
\end{figure}

\subsection{Round 3, Version D: Dual-Layer Guidance}

\paragraph{Motivation.}
The unexpected equivalence of A and B in Round~3, combined with C's
partial recovery, motivates the design of Version~D: if prompt-level
rules are effective but incomplete, and in-file metadata is complete
but loses effectiveness at scale, combining them should allow the
prompt rules to handle the cases that in-file metadata fails on, while
in-file metadata provides structural scaffolding for the cases where
the prompt rules do not name a specific category.

\paragraph{Results.}
Version~D achieves PA = 20/20 (100\%) in Round~3, recovering all
seven questions that Version~C still misses.
Secondary hit rate = 0/17 (0\%).
Total score = 20.0/28.5.

The seven recovered misses (Q02, Q12, Q13, Q14, Q15, Q16, Q18/Q19)
all correspond to the abstraction-level confusion pattern: in each case,
the Version~D prompt explicitly names the target category and states
that broader pipeline/governance categories should be preferred over
narrower mechanism categories.

\paragraph{Analysis.}
The perfect primary accuracy of Version~D confirms
Hypothesis~\ref{hyp:dual}: the two guidance layers are genuinely
complementary. Table~\ref{tab:miss_recovery} maps each question's
recovery status across conditions, showing which layer is responsible
for each recovery.

\begin{table}[t]
\centering
\caption{Question-level miss recovery across conditions in Round~3.
         \checkmark = correct, \xmark = incorrect. ``Recovery layer''
         identifies which element of Version~D resolves the miss.}
\label{tab:miss_recovery}
\small
\begin{tabular}{clccccl}
\toprule
\textbf{Q\#} & \textbf{Target Category} & \textbf{A} & \textbf{B} & \textbf{C} & \textbf{D} & \textbf{Recovery Layer} \\
\midrule
Q02 & Axiomatic\_Logic\_\&\_Audit\_Systems      & $\times$ & $\times$ & $\times$ & \checkmark & Explicit naming in prompt \\
Q12 & Distributed\_Cognition\_\&\_Context\_Orch.& $\times$ & $\times$ & $\times$ & \checkmark & Priority rule (governance $>$ mechanics) \\
Q13 & Adversarial\_Systems\_Thinking           & $\times$ & $\times$ & $\times$ & \checkmark & Priority rule (framework $>$ intelligence) \\
Q14 & Academic\_Research\_Synthesis\_Pipeline  & $\times$ & $\times$ & $\times$ & \checkmark & Priority rule (pipeline $>$ component) \\
Q15 & Revenue\_Generation\_\&\_Commercial\_Logic& $\times$ & $\times$ & \checkmark & \checkmark & Explicit naming in prompt (C, D) \\
Q16 & Cross\_Cultural\_Localization\_Intelligence& $\times$ & \checkmark& \checkmark & \checkmark & In-file routing\_hint (B, C, D) \\
Q18 & Interactive\_Narrative\_\&\_Fiction\_Engine& $\times$ & $\times$ & \checkmark & \checkmark & Explicit naming in prompt (C, D) \\
Q19 & Minimalist\_Entrepreneurship\_Execution  & $\times$ & $\times$ & \checkmark & \checkmark & Explicit naming in prompt (C, D) \\
\bottomrule
\end{tabular}
\end{table}

\subsection{Secondary Routing: A Persistent Failure}

Across all four conditions and all three rounds, secondary hit rates
remain consistently low (maximum 3/17 = 17.6\%, in Version~B Round~2).
Version~D achieves 0/17 secondary hits despite perfect primary accuracy.

Inspection of secondary routing errors reveals a consistent pattern:
the model selects a \emph{functionally adjacent} category within the
same semantic neighbourhood rather than an \emph{architecturally complementary}
category at a different abstraction level. For example:

\begin{itemize}
  \item Q05 (primary: \texttt{Persona\_\&\_Narrative\_Synthesis}):
    secondary answer key expects
    \texttt{Interactive\_Narrative\_\&\_Creative\_Fiction\_Engine}
    (execution layer); model consistently selects
    \texttt{Character\_Voice\_Synthesis} (same semantic neighbourhood).

  \item Q17 (primary: \texttt{Self\_Evolution\_\&\_Refinement}):
    answer key expects \texttt{Meta\_Data\_\&\_Engineering}
    (engineering foundation); model selects
    \texttt{Prompt\_Iteration\_Tracker} (same functional neighbourhood).
\end{itemize}

This failure is not addressable by either guidance layer because the
architectural complementarity relationship---the reason
\texttt{Meta\_Data\_\&\_Engineering} is the ``correct'' complement to
\texttt{Self\_Evolution\_\&\_Refinement}---is not encoded anywhere in the
current data structure. It exists only in the library designer's mental model.
We discuss the implications in Section~\ref{sec:analysis}.

\begin{figure}[t]
\centering
\begin{tikzpicture}
\begin{axis}[
  ybar,
  bar width=12pt,
  width=0.9\linewidth,
  height=6cm,
  ylabel={Count (out of max)},
  symbolic x coords={Primary hits (max 20), Secondary hits (max 17), Total score (max 28.5)},
  xtick=data,
  x tick label style={font=\small, align=center, text width=3cm},
  ymin=0, ymax=25,
  legend style={at={(0.5,1.05)}, anchor=south, legend columns=4, font=\small},
  nodes near coords,
  nodes near coords align={vertical},
  every node near coord/.append style={font=\footnotesize},
  grid=major,
  grid style={dashed,gray!30},
]
\addplot[fill=verA, fill opacity=0.85] coordinates
  {(Primary hits (max 20),13) (Secondary hits (max 17),0) (Total score (max 28.5),13)};
\addlegendentry{A}

\addplot[fill=verB, fill opacity=0.85] coordinates
  {(Primary hits (max 20),13) (Secondary hits (max 17),0) (Total score (max 28.5),13)};
\addlegendentry{B}

\addplot[fill=verC, fill opacity=0.85] coordinates
  {(Primary hits (max 20),16) (Secondary hits (max 17),1) (Total score (max 28.5),16.5)};
\addlegendentry{C}

\addplot[fill=verD, fill opacity=0.85] coordinates
  {(Primary hits (max 20),20) (Secondary hits (max 17),0) (Total score (max 28.5),20)};
\addlegendentry{D}

\end{axis}
\end{tikzpicture}
\caption{Round~3 (119 categories) results by metric and condition.
         Version~D achieves perfect primary accuracy but zero secondary hits,
         revealing the fundamental asymmetry between primary and secondary routing.}
\label{fig:round3_bar}
\end{figure}
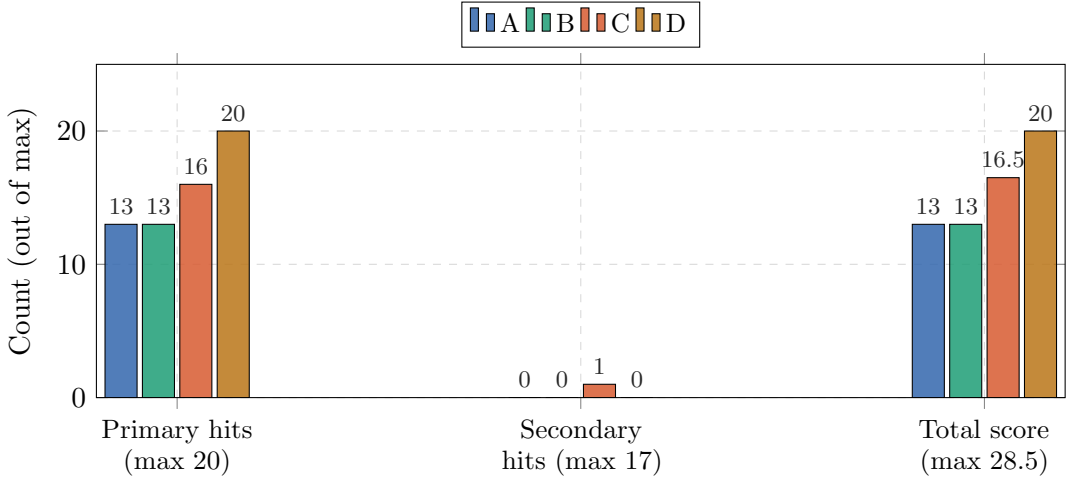

\begin{figure}[t]
\centering
\begin{tikzpicture}[font=\scriptsize, xscale=1.0]
\colorlet{hitcol}{verB!35}
\colorlet{misscol}{verC!40}
\node[font=\small\bfseries] at (1.35, 10.55) {A};
\node[font=\small\bfseries] at (2.15, 10.55) {B};
\node[font=\small\bfseries] at (2.95, 10.55) {C};
\node[font=\small\bfseries] at (3.75, 10.55) {D};
\fill[hitcol] (1.00,-0.50) rectangle (1.70,-0.15);
\node[anchor=west,font=\scriptsize] at (1.75,-0.33) {Correct};
\fill[misscol] (2.60,-0.50) rectangle (3.30,-0.15);
\node[anchor=west,font=\scriptsize] at (3.35,-0.33) {Incorrect};
\node[anchor=east,font=\scriptsize] at (0.90, 0.43) {Q20};
\fill[hitcol] (1.00,0.10) rectangle (1.70,0.75); \node at (1.35,0.43) {\textcolor{verB!80!black}{\checkmark}};
\fill[hitcol] (1.80,0.10) rectangle (2.50,0.75); \node at (2.15,0.43) {\textcolor{verB!80!black}{\checkmark}};
\fill[hitcol] (2.60,0.10) rectangle (3.30,0.75); \node at (2.95,0.43) {\textcolor{verB!80!black}{\checkmark}};
\fill[hitcol] (3.40,0.10) rectangle (4.10,0.75); \node at (3.75,0.43) {\textcolor{verB!80!black}{\checkmark}};
\node[anchor=east,font=\scriptsize] at (0.90, 0.93) {Q19};
\fill[misscol] (1.00,0.60) rectangle (1.70,1.25); \node at (1.35,0.93) {\textcolor{verC!80!black}{$\times$}};
\fill[misscol] (1.80,0.60) rectangle (2.50,1.25); \node at (2.15,0.93) {\textcolor{verC!80!black}{$\times$}};
\fill[hitcol] (2.60,0.60) rectangle (3.30,1.25); \node at (2.95,0.93) {\textcolor{verB!80!black}{\checkmark}};
\fill[hitcol] (3.40,0.60) rectangle (4.10,1.25); \node at (3.75,0.93) {\textcolor{verB!80!black}{\checkmark}};
\node[anchor=east,font=\scriptsize] at (0.90, 1.43) {Q18};
\fill[misscol] (1.00,1.10) rectangle (1.70,1.75); \node at (1.35,1.43) {\textcolor{verC!80!black}{$\times$}};
\fill[misscol] (1.80,1.10) rectangle (2.50,1.75); \node at (2.15,1.43) {\textcolor{verC!80!black}{$\times$}};
\fill[hitcol] (2.60,1.10) rectangle (3.30,1.75); \node at (2.95,1.43) {\textcolor{verB!80!black}{\checkmark}};
\fill[hitcol] (3.40,1.10) rectangle (4.10,1.75); \node at (3.75,1.43) {\textcolor{verB!80!black}{\checkmark}};
\node[anchor=east,font=\scriptsize] at (0.90, 1.93) {Q17};
\fill[hitcol] (1.00,1.60) rectangle (1.70,2.25); \node at (1.35,1.93) {\textcolor{verB!80!black}{\checkmark}};
\fill[hitcol] (1.80,1.60) rectangle (2.50,2.25); \node at (2.15,1.93) {\textcolor{verB!80!black}{\checkmark}};
\fill[hitcol] (2.60,1.60) rectangle (3.30,2.25); \node at (2.95,1.93) {\textcolor{verB!80!black}{\checkmark}};
\fill[hitcol] (3.40,1.60) rectangle (4.10,2.25); \node at (3.75,1.93) {\textcolor{verB!80!black}{\checkmark}};
\node[anchor=east,font=\scriptsize] at (0.90, 2.43) {Q16};
\fill[misscol] (1.00,2.10) rectangle (1.70,2.75); \node at (1.35,2.43) {\textcolor{verC!80!black}{$\times$}};
\fill[hitcol] (1.80,2.10) rectangle (2.50,2.75); \node at (2.15,2.43) {\textcolor{verB!80!black}{\checkmark}};
\fill[hitcol] (2.60,2.10) rectangle (3.30,2.75); \node at (2.95,2.43) {\textcolor{verB!80!black}{\checkmark}};
\fill[hitcol] (3.40,2.10) rectangle (4.10,2.75); \node at (3.75,2.43) {\textcolor{verB!80!black}{\checkmark}};
\node[anchor=east,font=\scriptsize] at (0.90, 2.93) {Q15};
\fill[misscol] (1.00,2.60) rectangle (1.70,3.25); \node at (1.35,2.93) {\textcolor{verC!80!black}{$\times$}};
\fill[misscol] (1.80,2.60) rectangle (2.50,3.25); \node at (2.15,2.93) {\textcolor{verC!80!black}{$\times$}};
\fill[hitcol] (2.60,2.60) rectangle (3.30,3.25); \node at (2.95,2.93) {\textcolor{verB!80!black}{\checkmark}};
\fill[hitcol] (3.40,2.60) rectangle (4.10,3.25); \node at (3.75,2.93) {\textcolor{verB!80!black}{\checkmark}};
\node[anchor=east,font=\scriptsize] at (0.90, 3.43) {Q14};
\fill[misscol] (1.00,3.10) rectangle (1.70,3.75); \node at (1.35,3.43) {\textcolor{verC!80!black}{$\times$}};
\fill[misscol] (1.80,3.10) rectangle (2.50,3.75); \node at (2.15,3.43) {\textcolor{verC!80!black}{$\times$}};
\fill[misscol] (2.60,3.10) rectangle (3.30,3.75); \node at (2.95,3.43) {\textcolor{verC!80!black}{$\times$}};
\fill[hitcol] (3.40,3.10) rectangle (4.10,3.75); \node at (3.75,3.43) {\textcolor{verB!80!black}{\checkmark}};
\node[anchor=east,font=\scriptsize] at (0.90, 3.93) {Q13};
\fill[misscol] (1.00,3.60) rectangle (1.70,4.25); \node at (1.35,3.93) {\textcolor{verC!80!black}{$\times$}};
\fill[misscol] (1.80,3.60) rectangle (2.50,4.25); \node at (2.15,3.93) {\textcolor{verC!80!black}{$\times$}};
\fill[misscol] (2.60,3.60) rectangle (3.30,4.25); \node at (2.95,3.93) {\textcolor{verC!80!black}{$\times$}};
\fill[hitcol] (3.40,3.60) rectangle (4.10,4.25); \node at (3.75,3.93) {\textcolor{verB!80!black}{\checkmark}};
\node[anchor=east,font=\scriptsize] at (0.90, 4.43) {Q12};
\fill[misscol] (1.00,4.10) rectangle (1.70,4.75); \node at (1.35,4.43) {\textcolor{verC!80!black}{$\times$}};
\fill[misscol] (1.80,4.10) rectangle (2.50,4.75); \node at (2.15,4.43) {\textcolor{verC!80!black}{$\times$}};
\fill[misscol] (2.60,4.10) rectangle (3.30,4.75); \node at (2.95,4.43) {\textcolor{verC!80!black}{$\times$}};
\fill[hitcol] (3.40,4.10) rectangle (4.10,4.75); \node at (3.75,4.43) {\textcolor{verB!80!black}{\checkmark}};
\node[anchor=east,font=\scriptsize] at (0.90, 4.93) {Q11};
\fill[hitcol] (1.00,4.60) rectangle (1.70,5.25); \node at (1.35,4.93) {\textcolor{verB!80!black}{\checkmark}};
\fill[hitcol] (1.80,4.60) rectangle (2.50,5.25); \node at (2.15,4.93) {\textcolor{verB!80!black}{\checkmark}};
\fill[hitcol] (2.60,4.60) rectangle (3.30,5.25); \node at (2.95,4.93) {\textcolor{verB!80!black}{\checkmark}};
\fill[hitcol] (3.40,4.60) rectangle (4.10,5.25); \node at (3.75,4.93) {\textcolor{verB!80!black}{\checkmark}};
\node[anchor=east,font=\scriptsize] at (0.90, 5.43) {Q10};
\fill[hitcol] (1.00,5.10) rectangle (1.70,5.75); \node at (1.35,5.43) {\textcolor{verB!80!black}{\checkmark}};
\fill[hitcol] (1.80,5.10) rectangle (2.50,5.75); \node at (2.15,5.43) {\textcolor{verB!80!black}{\checkmark}};
\fill[hitcol] (2.60,5.10) rectangle (3.30,5.75); \node at (2.95,5.43) {\textcolor{verB!80!black}{\checkmark}};
\fill[hitcol] (3.40,5.10) rectangle (4.10,5.75); \node at (3.75,5.43) {\textcolor{verB!80!black}{\checkmark}};
\node[anchor=east,font=\scriptsize] at (0.90, 5.93) {Q09};
\fill[hitcol] (1.00,5.60) rectangle (1.70,6.25); \node at (1.35,5.93) {\textcolor{verB!80!black}{\checkmark}};
\fill[hitcol] (1.80,5.60) rectangle (2.50,6.25); \node at (2.15,5.93) {\textcolor{verB!80!black}{\checkmark}};
\fill[hitcol] (2.60,5.60) rectangle (3.30,6.25); \node at (2.95,5.93) {\textcolor{verB!80!black}{\checkmark}};
\fill[hitcol] (3.40,5.60) rectangle (4.10,6.25); \node at (3.75,5.93) {\textcolor{verB!80!black}{\checkmark}};
\node[anchor=east,font=\scriptsize] at (0.90, 6.43) {Q08};
\fill[hitcol] (1.00,6.10) rectangle (1.70,6.75); \node at (1.35,6.43) {\textcolor{verB!80!black}{\checkmark}};
\fill[hitcol] (1.80,6.10) rectangle (2.50,6.75); \node at (2.15,6.43) {\textcolor{verB!80!black}{\checkmark}};
\fill[hitcol] (2.60,6.10) rectangle (3.30,6.75); \node at (2.95,6.43) {\textcolor{verB!80!black}{\checkmark}};
\fill[hitcol] (3.40,6.10) rectangle (4.10,6.75); \node at (3.75,6.43) {\textcolor{verB!80!black}{\checkmark}};
\node[anchor=east,font=\scriptsize] at (0.90, 6.93) {Q07};
\fill[hitcol] (1.00,6.60) rectangle (1.70,7.25); \node at (1.35,6.93) {\textcolor{verB!80!black}{\checkmark}};
\fill[hitcol] (1.80,6.60) rectangle (2.50,7.25); \node at (2.15,6.93) {\textcolor{verB!80!black}{\checkmark}};
\fill[hitcol] (2.60,6.60) rectangle (3.30,7.25); \node at (2.95,6.93) {\textcolor{verB!80!black}{\checkmark}};
\fill[hitcol] (3.40,6.60) rectangle (4.10,7.25); \node at (3.75,6.93) {\textcolor{verB!80!black}{\checkmark}};
\node[anchor=east,font=\scriptsize] at (0.90, 7.43) {Q06};
\fill[hitcol] (1.00,7.10) rectangle (1.70,7.75); \node at (1.35,7.43) {\textcolor{verB!80!black}{\checkmark}};
\fill[hitcol] (1.80,7.10) rectangle (2.50,7.75); \node at (2.15,7.43) {\textcolor{verB!80!black}{\checkmark}};
\fill[hitcol] (2.60,7.10) rectangle (3.30,7.75); \node at (2.95,7.43) {\textcolor{verB!80!black}{\checkmark}};
\fill[hitcol] (3.40,7.10) rectangle (4.10,7.75); \node at (3.75,7.43) {\textcolor{verB!80!black}{\checkmark}};
\node[anchor=east,font=\scriptsize] at (0.90, 7.93) {Q05};
\fill[hitcol] (1.00,7.60) rectangle (1.70,8.25); \node at (1.35,7.93) {\textcolor{verB!80!black}{\checkmark}};
\fill[hitcol] (1.80,7.60) rectangle (2.50,8.25); \node at (2.15,7.93) {\textcolor{verB!80!black}{\checkmark}};
\fill[hitcol] (2.60,7.60) rectangle (3.30,8.25); \node at (2.95,7.93) {\textcolor{verB!80!black}{\checkmark}};
\fill[hitcol] (3.40,7.60) rectangle (4.10,8.25); \node at (3.75,7.93) {\textcolor{verB!80!black}{\checkmark}};
\node[anchor=east,font=\scriptsize] at (0.90, 8.43) {Q04};
\fill[hitcol] (1.00,8.10) rectangle (1.70,8.75); \node at (1.35,8.43) {\textcolor{verB!80!black}{\checkmark}};
\fill[hitcol] (1.80,8.10) rectangle (2.50,8.75); \node at (2.15,8.43) {\textcolor{verB!80!black}{\checkmark}};
\fill[hitcol] (2.60,8.10) rectangle (3.30,8.75); \node at (2.95,8.43) {\textcolor{verB!80!black}{\checkmark}};
\fill[hitcol] (3.40,8.10) rectangle (4.10,8.75); \node at (3.75,8.43) {\textcolor{verB!80!black}{\checkmark}};
\node[anchor=east,font=\scriptsize] at (0.90, 8.93) {Q03};
\fill[hitcol] (1.00,8.60) rectangle (1.70,9.25); \node at (1.35,8.93) {\textcolor{verB!80!black}{\checkmark}};
\fill[hitcol] (1.80,8.60) rectangle (2.50,9.25); \node at (2.15,8.93) {\textcolor{verB!80!black}{\checkmark}};
\fill[hitcol] (2.60,8.60) rectangle (3.30,9.25); \node at (2.95,8.93) {\textcolor{verB!80!black}{\checkmark}};
\fill[hitcol] (3.40,8.60) rectangle (4.10,9.25); \node at (3.75,8.93) {\textcolor{verB!80!black}{\checkmark}};
\node[anchor=east,font=\scriptsize] at (0.90, 9.43) {Q02};
\fill[misscol] (1.00,9.10) rectangle (1.70,9.75); \node at (1.35,9.43) {\textcolor{verC!80!black}{$\times$}};
\fill[misscol] (1.80,9.10) rectangle (2.50,9.75); \node at (2.15,9.43) {\textcolor{verC!80!black}{$\times$}};
\fill[misscol] (2.60,9.10) rectangle (3.30,9.75); \node at (2.95,9.43) {\textcolor{verC!80!black}{$\times$}};
\fill[hitcol] (3.40,9.10) rectangle (4.10,9.75); \node at (3.75,9.43) {\textcolor{verB!80!black}{\checkmark}};
\node[anchor=east,font=\scriptsize] at (0.90, 9.93) {Q01};
\fill[hitcol] (1.00,9.60) rectangle (1.70,10.25); \node at (1.35,9.93) {\textcolor{verB!80!black}{\checkmark}};
\fill[hitcol] (1.80,9.60) rectangle (2.50,10.25); \node at (2.15,9.93) {\textcolor{verB!80!black}{\checkmark}};
\fill[hitcol] (2.60,9.60) rectangle (3.30,10.25); \node at (2.95,9.93) {\textcolor{verB!80!black}{\checkmark}};
\fill[hitcol] (3.40,9.60) rectangle (4.10,10.25); \node at (3.75,9.93) {\textcolor{verB!80!black}{\checkmark}};
\end{tikzpicture}
\caption{Per-question primary routing hit/miss matrix for Round~3 (119 categories).
         Green \checkmark = correct primary category; red $\times$ = incorrect.
         Questions Q02, Q12, Q13, Q14 are missed by all conditions except~D,
         indicating persistent abstraction-level confusion unresolvable by
         in-file guidance or simple prompt hints.}
\label{fig:heatmap}
\end{figure}

\section{Analysis and Discussion}
\label{sec:analysis}

\subsection{The Primary/Secondary Routing Asymmetry}

Our experiments reveal a fundamental asymmetry between two routing tasks
that superficially appear similar. \emph{Primary routing} is a
classification problem: given a task description and a library, select the
single most relevant category. This problem is solvable because (i)~each
category has a unique name that, when explicitly pointed to in a prompt,
can be matched to a task description, and (ii)~an abstract priority rule
(``prefer pipeline over component'') eliminates the systematic
abstraction-level confusion.

\emph{Secondary routing} is qualitatively different. The answer key's
secondary categories are not the second-most-similar category in surface
keyword space---they are categories that address a \emph{different
architectural dimension} of the same task. The relationship
$R_2(\text{Q05}) = \texttt{Interactive\_Narrative}$ while
$R_1(\text{Q05}) = \texttt{Persona\_\&\_Narrative}$ reflects a design
decision by the library architect: narratives require both character
consistency (Persona layer) and plot execution infrastructure
(Interactive Narrative layer). This relationship is not inferable from
any text currently in the library.

\begin{principle}[Architectural Intent Encoding]
\label{princ:intent}
Cross-category complementarity relationships that form part of a
knowledge library's intended usage pattern must be explicitly encoded
in the data structure (e.g., a \texttt{complement} field on each
category) rather than expected to be inferred by the LLM from
category descriptions. LLMs resolve secondary selection by surface
proximity; intended complementarity is invisible without explicit encoding.
\end{principle}

\paragraph{Why prompt instructions cannot substitute for explicit encoding.}
Our experimental data provide a precise characterisation of this boundary.
Across all four guidance conditions (A, B, C, D) and all three library
scales (36, 60, and 119 categories), secondary hit rates remain at or near
zero---with the sole partial exception of Version~B Round~2 (3/17 = 17.6\%),
which is not statistically meaningful given the single-run design.
Critically, this failure is \emph{scale-independent} and
\emph{instruction-independent}: it persists regardless of library size,
and regardless of whether the system prompt contains explicit routing rules.
Version~D, which achieves perfect primary accuracy (20/20) through explicit
category naming and priority rules, still achieves 0/17 secondary hits.
This dissociation confirms that the failure is not a symptom of attentional
overload or insufficient guidance---it is a knowledge absence.

The underlying mechanism can be stated precisely. LLMs acquire knowledge
through statistical co-occurrence in training data. The concept of
\emph{architectural complementarity} is present in that training
data---LLMs have encountered countless descriptions of how execution layers
depend on engineering foundations, how governance layers constrain mechanism
layers, and so on. However, the \emph{specific pairings} defined by a
practitioner's private library design---that
\texttt{Self\_Evolution\_\&\_Refinement} pairs with
\texttt{Meta\_Data\_\&\_Engineering}, or that
\texttt{Persona\_\&\_Narrative\_Synthesis} pairs with
\texttt{Interactive\_Narrative\_\&\_Creative\_Fiction\_Engine}---have never
appeared in any training corpus, because they are the products of a single
designer's architectural decisions. This exposes a boundary in LLM capability
that is easy to overlook: the distinction between \emph{knowing a concept}
and \emph{knowing its instantiation in a private context}.

Furthermore, ``architectural complementarity'' and ``semantic similarity''
point in \emph{opposite directions} in embedding space: a category's
architectural complement is typically drawn from a different abstraction
level and a different functional domain, making it semantically
\emph{distant} rather than proximate. An LLM's default retrieval
behaviour---select the nearest semantic neighbour---is therefore
systematically misaligned with the secondary routing task, regardless
of how the task is framed in the prompt. The practical implication is
a hard design constraint: \emph{any knowledge that exists only in the
library designer's mental model must be explicitly serialised into the
data structure before an LLM can act on it.}

\subsection{The Scalability Boundary of In-File Guidance}

Version~B's collapse from PA=16 (Round~2) to PA=13 (Round~3)---identical
to the no-guidance baseline---reveals a concrete scalability boundary
for in-file routing hints. The 100-character routing hints collectively
occupy approximately 12,000 tokens in the \texttt{category\_index} at
119 categories. This block, while positioned at the file's primacy location,
is itself long enough to suffer internal lost-in-the-middle degradation.
Entries near the middle of the 119-entry index receive weaker attention
than those at the beginning and end.

This suggests a practical design rule:

\begin{principle}[Routing Hint Density Limit]
\label{princ:density}
In-file routing hint indexes are effective up to approximately 60
categories when using 100-character hints. Beyond this scale, the index
block itself becomes too long for reliable end-to-end attention. Solutions
include: (i)~shortening hints to 50 characters, (ii)~splitting the
library into multiple files with a meta-index, or (iii)~adding
explicit prompt-level priority rules for the most frequently confused
categories (the Dual-Layer approach).
\end{principle}

\subsection{Three Principles of Effective Prompt-Level Guidance}

Version~C and~D's success relative to~B at large scale, combined with~D's
perfect accuracy, allows us to distil three principles for prompt-level
guidance design:

\begin{principle}[Explicit Category Naming]
\label{princ:naming}
For categories that are frequently confused with semantically proximate
distractors, explicit naming in the system prompt (``For multi-agent
workflow tasks, the correct category is
\texttt{Distributed\_Cognition\_\&\_Context\_Orchestration}'')
is more reliable than relying on the model to navigate a long index.
Generic structural instructions without specific names are insufficient.
\end{principle}

\begin{principle}[Abstract Hierarchy Rule]
\label{princ:hierarchy}
A single abstract priority rule---``prefer broader pipeline/governance
categories over narrower mechanism/component categories when both seem
relevant''---resolves the most common class of routing errors at scale.
This rule addresses the pattern where a distractor name is more literally
descriptive of the task's surface mechanics while the correct category
addresses the task's systemic function.
\end{principle}

\begin{principle}[Complementary Layer Division of Labour]
\label{princ:complementary}
Effective dual-layer guidance requires that the two layers address
\emph{different} routing challenges: in-file metadata handles structural
navigation over the full category space, while prompt rules handle
abstraction-level disambiguation for specific high-risk category pairs.
Duplicate content across layers provides no additional benefit.
\end{principle}

\subsection{Cognitive Framing: Data Object vs.\ Instruction Context}

A striking empirical finding is that navigational instructions embedded
\emph{within a data file} and identical instructions provided in the
\emph{system prompt} produce different routing behaviour, even when both
occupy primacy positions within their respective contexts. This observation
is consistent with the primacy bias documented by \citet{liu2024lost},
who demonstrate that LLMs attend most strongly to tokens near the beginning
of their input context; instructions co-located with the data they describe
benefit from this bias within the data object's own processing scope.

We hypothesise that this reflects a difference in cognitive framing:
the LLM processes the system prompt as an instructional context
establishing its role and operating procedures, while it processes the
uploaded data file as a content object to be navigated. Instructions
embedded within the data object benefit from being co-processed with the
data they describe---a form of inline contextualisation---whereas system
prompt instructions must bridge a representational gap to influence
navigation of a separately-encoded data object.

This framing predicts that at small scales (Round~1), both placements
are equally effective because the navigational challenge is trivial.
At medium scales (Round~2), in-file guidance provides a modest advantage
via primacy bias within the data object. At large scales (Round~3),
the advantage reverses because the data object's internal primacy-bias
effect degrades over the 119-entry index, while the system prompt's
abstract rules maintain full effectiveness regardless of library size.

\subsection{Limitations}

Several limitations constrain the generalisability of our findings:

\begin{enumerate}
  \item \textbf{Single model and single run}: All experiments use
    Claude Opus~4.6 with a single run per condition. LLM responses
    have non-zero temperature, introducing sampling variance.
    Future work should replicate with multiple runs (at least $n=3$)
    and additional models (GPT-4o, Gemini).

  \item \textbf{Human-designed answer key}: The secondary routing
    answer key reflects the library designer's architectural intent,
    not independently validated ground truth. A different expert might
    assign different secondary categories to some questions.

  \item \textbf{Single knowledge domain}: HISL is a prompt-engineering
    skill library. Generalisation to other domains (medical, legal,
    scientific) requires additional validation, though the mechanisms
    we identify---positional bias, abstraction-level confusion,
    surface keyword proximity---are domain-independent.

  \item \textbf{Q02, Q12, Q13, Q14 structural issues}: The four
    permanently-lost questions reflect a mismatch between the distractor
    design and the answer key's abstraction-level preferences. These
    may reflect answer key issues as much as model limitations.
    \item \textbf{Architectural dependence on autoregressive primacy bias}:
    SDSR's core mechanism---placing navigational metadata at the file's
    primacy position---presupposes that the LLM exhibits primacy bias.
    \citet{wu2025emergence} show that this bias is a structural
    consequence of causal attention masking, and
    \citet{salvatore2025lost} demonstrate that the primacy effect
    emerges in autoregressive architectures but not in bidirectional
    encoder-decoders. If future mainstream LLM architectures adopt
    bidirectional or modified masking schemes---as proposed by
    \citet{wu2025emergence} as a mitigation strategy---the
    primacy-exploitation component of SDSR would need to be
    re-evaluated. The prompt-level guidance layer (Version~C/D),
    however, operates through explicit instruction-following rather
    than positional bias, and is therefore expected to remain effective
    across architectural changes.
\end{enumerate}

\section{Self-Describing Structured Retrieval as a RAG Alternative}
\label{sec:sdsr}

\subsection{RAG's Fundamental Assumptions and Their Costs}

Retrieval-Augmented Generation \citep{lewis2020rag} makes three implicit
assumptions that are often violated in expert-curated knowledge libraries:

\begin{enumerate}
  \item \textbf{Surface similarity $\approx$ relevance}: Vector similarity
    retrieval assumes that semantically relevant documents will be close in
    embedding space to the query. For human-defined categories whose
    distinctions are definitional rather than distributional (e.g.,
    ``governance layer'' vs.\ ``execution layer''), this assumption fails.

  \item \textbf{Chunks are self-contained}: Fixed-size chunking assumes
    that relevant information can be extracted from a document fragment.
    For structured libraries, a skill's meaning depends on its category
    membership; splitting them severs this dependency.

  \item \textbf{Boundaries are not pre-known}: RAG assumes that the
    relevant scope of a query cannot be determined without vector search.
    For curated libraries, the designer already knows the category
    boundaries; the retrieval problem is routing, not discovery.
\end{enumerate}

Beyond these assumptions, RAG carries operational costs that are unnecessary
for structured libraries: embedding model inference, vector database
infrastructure (FAISS, Weaviate, Pinecone), index maintenance on updates,
and chunking strategy tuning. \citet{packowski2024optimizing} report
that enterprise RAG deployments require iterative content design and
evaluation effort to achieve stable results, with significant total
cost of ownership.

\subsection{The SDSR Architecture}

\begin{definition}[Self-Describing Structured Retrieval]
\emph{Self-Describing Structured Retrieval (SDSR)} is a two-tier
retrieval architecture in which:
(1)~a structured knowledge file embeds a \texttt{\_summary} block
containing a human-authored index of its categories, positioned at
the file's primacy location;
(2)~a Python orchestration layer reads only the \texttt{\_summary}
block of each file to identify relevant files, then loads only the
targeted files' full content into the LLM context.
\end{definition}

\begin{figure}[h]
\centering
\begin{tikzpicture}[
node distance=0.5cm and 0.6cm,
  box/.style={rectangle, rounded corners=4pt, draw, minimum width=2.2cm,
              minimum height=0.8cm, align=center, font=\small},
  decision/.style={diamond, draw, aspect=0.8
  , align=center, font=\small},
  arrow/.style={-Stealth, thick},
  highlight/.style={box, fill=verB!20, draw=verB},
  normal/.style={box, fill=gray!10},
  python/.style={box, fill=verC!20, draw=verC},
]

\node[normal] (query) {User query $q$};
\node[python, right=of query] (sumread) {Python reads\\$\texttt{\_summary}$ only\\(all files)};
\node[decision, right=of sumread] (route) {Relevant\\files?};
\node[highlight, right=of route] (fullload) {Load full content\\of matched files};
\node[python, right=of fullload] (llm) {LLM processes\\(small context)};
\node[normal, right=of llm] (ans) {Answer + \\skill selection};

\draw[arrow] (query) -- (sumread);
\draw[arrow] (sumread) -- (route);
\draw[arrow] (route) -- node[above,font=\scriptsize]{yes} (fullload);
\draw[arrow] (route.south) -- ++(0,-0.5) -- node[below,font=\scriptsize]{no match} ++(2,0) |-
   node[right,font=\scriptsize]{expand scope} (sumread);
\draw[arrow] (fullload) -- (llm);
\draw[arrow] (llm) -- (ans);

\node[font=\scriptsize, below=0.15cm of sumread, align=center, color=verC]
  {$\sim$200 tokens/file};
\node[font=\scriptsize, below=0.15cm of fullload, align=center, color=verB]
  {1--3 files\\$\leq$5K tokens};

\end{tikzpicture}
\caption{The SDSR two-tier retrieval pipeline. Python reads only the
         \texttt{\_summary} blocks (low cost) to identify relevant files,
         then loads full content (small context) for LLM processing.
         No vector database or embedding model is required.}
\label{fig:sdsr_pipeline}
\end{figure}
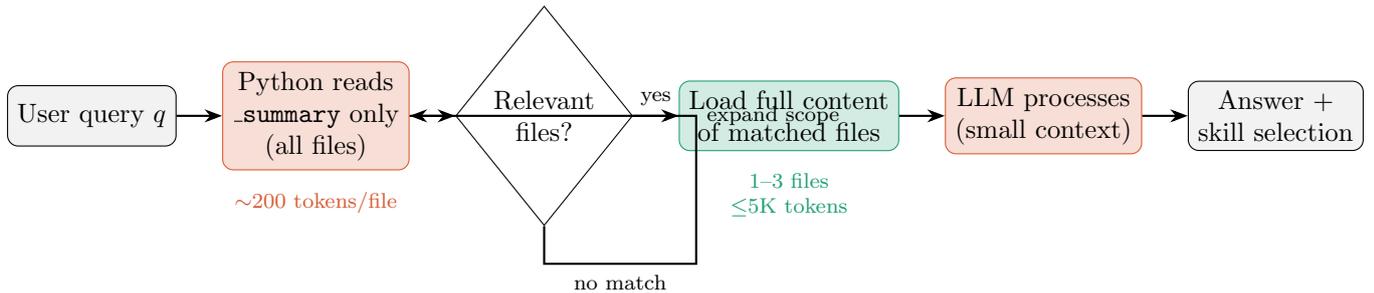

The SDSR pipeline (Figure~\ref{fig:sdsr_pipeline}) operates as follows:

\begin{algorithm}[t]
\caption{SDSR Two-Tier Retrieval}
\label{alg:sdsr}
\begin{algorithmic}[1]
\Require Query $q$, file registry $\mathcal{F} = \{f_1, \ldots, f_K\}$
\Ensure Selected skill set $\mathcal{S}^*$
\State \textbf{Tier 1 — Summary Scan:}
\For{each file $f_i \in \mathcal{F}$}
  \State $\sigma_i \leftarrow \text{read\_json\_prefix}(f_i, \texttt{\_summary})$
  \Comment{$\sim$200 tokens per file}
\EndFor
\State $\mathcal{F}^* \leftarrow \text{LLM\_route}(q, \{\sigma_1, \ldots, \sigma_K\})$
\Comment{Identify 1--3 relevant files}
\State \textbf{Tier 2 — Full Content Processing:}
\For{each file $f_i \in \mathcal{F}^*$}
  \State $\text{context}_i \leftarrow \text{read\_json\_full}(f_i)$
\EndFor
\State $\mathcal{S}^* \leftarrow \text{LLM\_select}(q, \{\text{context}_i : f_i \in \mathcal{F}^*\})$
\Return $\mathcal{S}^*$
\end{algorithmic}
\end{algorithm}

The key efficiency property is that Tier~1 reads only the \texttt{\_summary}
block of each file (approximately 200 tokens per file regardless of file size),
making the total Tier~1 context proportional to the number of files rather
than their cumulative content. For a corpus of 100 files each containing
10,000 tokens, the Tier~1 context is approximately 20,000 tokens---within
any modern LLM's context window---compared to 1,000,000 tokens for naive
full-corpus injection.

\subsection{Comparison with RAG}

\begin{table}[t]
\centering
\caption{Comparison of SDSR versus traditional RAG across key dimensions.}
\label{tab:rag_comparison}
\small
\begin{tabular}{p{0.26\linewidth} p{0.32\linewidth} p{0.32\linewidth}}
\toprule
\textbf{Dimension} & \textbf{Traditional RAG} & \textbf{SDSR} \\
\midrule
Suitable data type & Unstructured, no pre-known boundaries & Structured, human-defined boundaries \\
\midrule
Index type & Vector embeddings (learned) & Human-authored routing hints (explicit) \\
\midrule
Retrieval mechanism & Cosine similarity over embedding space & LLM reads \texttt{\_summary} blocks \\
\midrule
Infrastructure & Vector DB (FAISS/Weaviate/Pinecone) & File system + Python \\
\midrule
Pre-processing cost & High: chunk, embed, index (per update) & Low: author \texttt{\_summary} (once) \\
\midrule
Semantic boundary preservation & At risk (chunking may sever) & Guaranteed (whole-category loading) \\
\midrule
Cross-reference encoding & Implicit in embedding space & Explicit (e.g., \texttt{complement} field) \\
\midrule
Scalability ceiling & Millions of tokens & $\sim$100 files $\times$ 50K tokens/file \\
\midrule
Boundary accuracy & Statistical (embedding-dependent) & Human-guaranteed \\
\bottomrule
\end{tabular}
\end{table}

Table~\ref{tab:rag_comparison} summarises the trade-offs. SDSR is strictly
preferable to RAG when: (i)~the knowledge base has explicit human-defined
semantic boundaries; (ii)~the total corpus size is within the scalable
range ($\leq$ a few million tokens across all files); and (iii)~the
pre-processing investment in authoring \texttt{\_summary} blocks is
acceptable (typically one LLM-assisted pass per file).

RAG retains its advantage for truly unstructured corpora (news archives,
raw web crawls, heterogeneous document collections) where no natural
structure exists to anchor a \texttt{\_summary} block.

\subsection{Extension: Structured Pre-Processing of Unstructured Documents}

SDSR's scope can be extended to semi-structured document types through
a one-time structuring pass. We illustrate with legal judgments as
a concrete example.

A judicial decision has recoverable structure: a plaintiff's claims section,
a defendant's response section, the court's reasoning section, and a
dispositif (holding). Rule-based or LLM-assisted extraction can identify
these sections automatically for the majority of judgment formats.
Once sectioned, a \texttt{\_summary} block can be authored that includes:

\begin{itemize}
  \item Core factual claims (1--2 sentences per party)
  \item The court's central legal reasoning (2--3 sentences)
  \item \textbf{Cross-references}: ``The court's reasoning in Section~C
    directly cites the plaintiff's damages argument in Section~A, paragraphs
    4--7; load both sections when the query involves quantum of damages.''
\end{itemize}

The cross-reference field is the SDSR equivalent of the \texttt{complement}
field for knowledge libraries: it makes explicit the relationships between
document sections that would otherwise require the LLM to infer from
full-text reading. This enables the Tier~1 routing phase to identify not
only the relevant judgment file, but also which sections of that file
must be co-loaded to answer the query.

\begin{figure}[t]
\centering
\begin{tikzpicture}[
  box/.style={rectangle, rounded corners=3pt, draw, minimum width=3cm,
              minimum height=0.7cm, align=center, font=\small, fill=gray!8},
  sbox/.style={box, fill=verB!15, draw=verB},
  arrow/.style={-Stealth, thick},
  label/.style={font=\scriptsize, color=gray!60!black},
]

\node[box] (raw) {Raw judgment text\\(unstructured)};
\node[box, right=1.5cm of raw] (sections) {Sectioned judgment\\P / D / Court / Holding};
\node[sbox, right=1.5cm of sections] (summary) {\texttt{\_summary} block\\with cross-references};
\node[box, below=0.8cm of summary] (tier1) {Tier~1: Python reads\\summary (200 tokens)};
\node[sbox, below=0.8cm of tier1] (tier2) {Tier~2: Load relevant\\sections + cross-refs};
\node[box, right=1.5cm of tier2] (llm) {LLM answers\\query};

\draw[arrow] (raw) -- node[above,label,align=center]{Rule-based\\sectioning} (sections);
\draw[arrow] (sections) -- node[above,label,align=center]{LLM summarise\\(once)} (summary);
\draw[arrow] (summary) -- (tier1);
\draw[arrow] (tier1) -- node[right,label]{match} (tier2);
\draw[arrow] (tier2) -- (llm);

\end{tikzpicture}
\caption{SDSR applied to legal judgment retrieval. A one-time structuring
         pass converts unstructured text into an SDSR-compatible file.
         Cross-reference fields in the \texttt{\_summary} block enable
         co-loading of dependent sections during Tier~2 retrieval.}
\label{fig:legal_sdsr}
\end{figure}
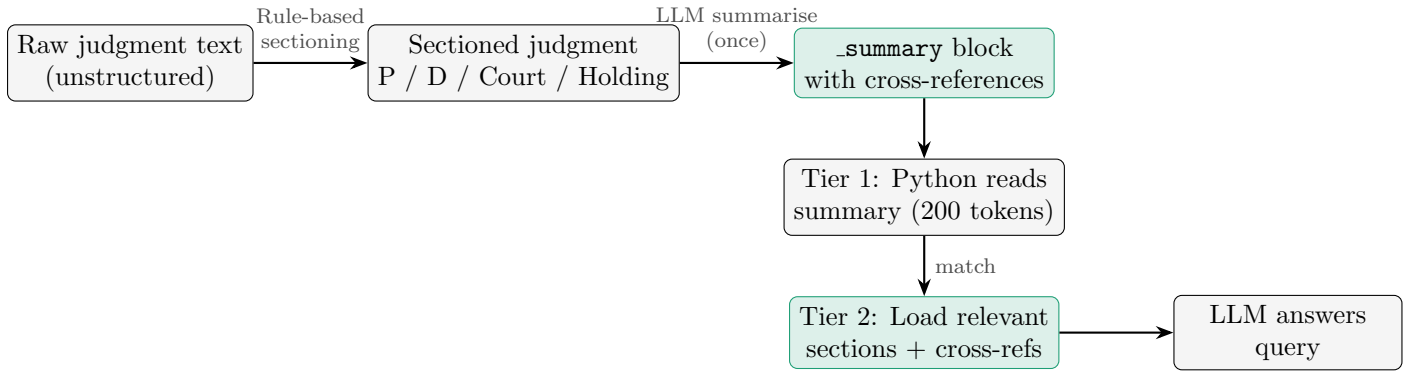

Pre-processing cost for this workflow is one LLM call per judgment to
generate the \texttt{\_summary} block, typically consuming 1,000--2,000
tokens. For a corpus of 10,000 judgments, the total pre-processing
cost is approximately 15--20 million tokens---comparable to the
embedding cost of a RAG pipeline over the same corpus, but producing
richer, explicitly-structured metadata rather than opaque dense vectors.

\section{Conclusion}
\label{sec:conclusion}

We have presented Self-Describing Structured Retrieval (SDSR), a framework
that exploits the LLM's positional primacy bias by embedding human-authored
navigational metadata at the beginning of structured data files. Through a
controlled four-round experiment with systematic category-count scaling and
adversarial distractor injection, we have established the following findings:

\begin{enumerate}
  \item \textbf{In-file guidance is effective up to $\sim$60 categories}
    but loses effectiveness at larger scales as the routing hint index
    itself becomes subject to lost-in-the-middle degradation.

  \item \textbf{Prompt-level guidance maintains effectiveness at scale}
    because abstract routing rules are fixed-length and occupy the system
    prompt's primacy position, independent of library size.

  \item \textbf{Dual-layer guidance (in-file + prompt) achieves perfect
    primary routing accuracy} (100\%, 20/20) at 119 categories,
    through complementary specialisation: in-file metadata handles
    structural navigation, prompt rules handle abstraction-level disambiguation.

  \item \textbf{Secondary routing requires explicit architectural intent
    encoding}: cross-category complementarity relationships are invisible
    to all guidance strategies unless encoded in a dedicated field
    (e.g., \texttt{complement}).

  \item \textbf{SDSR is a viable lightweight alternative to RAG} for
    structured knowledge libraries with human-defined semantic boundaries,
    eliminating the need for embedding models, vector databases, and
    chunking infrastructure while providing higher boundary accuracy.
\end{enumerate}

These findings have immediate practical implications for the design of
LLM-integrated knowledge management systems. Knowledge libraries should:
(i)~include a \texttt{\_summary} block with compact \texttt{routing\_hint}
fields; (ii)~pair this with a system prompt that names high-risk category
pairs and provides an abstract hierarchy rule; and (iii)~add explicit
\texttt{complement} fields to encode the architectural pairings that the
library designer intends.

\paragraph{Future Work.}
Priority directions include: (i)~multi-run replication with temperature
sampling to quantify variance; (ii)~cross-model generalisation to GPT-4o
and Gemini; (iii)~empirical validation of the \texttt{complement} field
approach to secondary routing; (iv)~SDSR implementation and benchmarking
for legal judgment retrieval; and (v)~investigation of the optimal
routing hint length as a function of category count; (vi)~integration with white-box position-bias mitigation methods
such as Ms-PoE \citep{zhang2024mspoe}, which rescales RoPE with
head-wise factors, or PINE \citep{wang2025pine}, which replaces
causal attention with bidirectional attention at the document level.
These model-level interventions address the same lost-in-the-middle
degradation that limits Version~B's scalability (Principle~\ref{princ:density}),
and could raise the effective category ceiling of in-file routing hints
from ${\sim}60$ to substantially higher counts. Evaluating whether
such white-box methods compose additively or synergistically with
SDSR's black-box dual-layer guidance would clarify a practical
deployment spectrum: pure black-box SDSR for API-only access,
SDSR~$+$~Ms-PoE/PINE for open-weight deployments, and
training-time interventions (e.g., FILM \citep{an2024film}) for
maximum debiasing at highest cost.

\clearpage
\bibliographystyle{unsrtnat}
\bibliography{references}

\clearpage
\appendix

\section{Full Test Question Set with Answer Keys}
\label{app:questions}

\begin{longtable}{p{0.04\linewidth} p{0.48\linewidth} p{0.25\linewidth} p{0.14\linewidth}}
\caption{Complete 20-question test set with primary and secondary target categories.
         Questions are designed with keyword avoidance: no question contains a substring
         of its primary target category name.}
\label{tab:full_questions} \\
\toprule
\textbf{Q\#} & \textbf{Task Description} & \textbf{Primary Target} & \textbf{Secondary Target} \\
\midrule
\endfirsthead
\multicolumn{4}{c}{{\tablename\ \thetable{} -- continued from previous page}}\\
\toprule
\textbf{Q\#} & \textbf{Task Description} & \textbf{Primary Target} & \textbf{Secondary Target} \\
\midrule
\endhead
\midrule
\multicolumn{4}{r}{{Continued on next page\ldots}}\\
\endfoot
\bottomrule
\endlastfoot
Q01 & ``I need to design an AI assistant that, upon receiving user input, can automatically determine what type of request it is and decide what processing strategy to apply.'' & \seqsplit{Cognitive\_Architecture\_\&\_Routing} & --- \\
Q02 & ``Help me build an AI system that can rigorously check its own output for logical contradictions and audit its own reasoning process step by step.'' & \seqsplit{Axiomatic\_Logic\_\&\_Audit\_Systems} & \seqsplit{Cognitive\_Architecture} \\
Q03 & ``I want an AI that, after reading an academic paper, can identify the unstated assumptions the author relies on and surface potential methodological weaknesses.'' & \seqsplit{Academic\_Insight\_\&\_Forensics} & \seqsplit{Axiomatic\_Logic} \\
Q04 & ``Design an interactive system for teaching elementary school students mathematics---one that dynamically adjusts difficulty based on student responses and makes the learning experience feel like a game.'' & \seqsplit{Interactive\_Pedagogy\_\&\_Gamification} & \seqsplit{Game\_Design\_\&\_Mechanics} \\
Q05 & ``I need an AI to play a detective character with a complex backstory, maintaining a consistent personality and speaking style throughout interactions with users.'' & \seqsplit{Persona\_\&\_Narrative\_Synthesis} & \seqsplit{Interactive\_Narrative\_Engine} \\
Q06 & ``Help me design a prompt system that can automatically extract structured data from messy unstructured input and convert it into a machine-readable format.'' & \seqsplit{Data\_Structuring\_\&\_Engineering} & \seqsplit{Meta\_Data\_\&\_Engineering} \\
Q07 & ``I'm developing an RPG game and need an AI to help design a combat balance system, including character progression curves and equipment power scaling formulas.'' & \seqsplit{Game\_Design\_\&\_Mechanics} & \seqsplit{RPG\_Narrative\_Director} \\
Q08 & ``Generate a web design specification that starts from a brand's visual identity and extends all the way to concrete CSS implementation.'' & \seqsplit{UI\_UX\_\&\_Frontend\_Engineering} & \seqsplit{Visual\_Architecture} \\
Q09 & ``I need to evaluate the potential social side-effects of a new policy, including indirect impacts and long-term external costs that are difficult to quantify.'' & \seqsplit{Policy\_Impact\_\&\_Externalities} & \seqsplit{Strategic\_Decision\_Frameworks} \\
Q10 & ``Design an AI system to act as a tabletop RPG game master---managing multiple players' actions, dice outcomes, and generating real-time narrative responses consistent with the world's lore.'' & \seqsplit{RPG\_Narrative\_Director} & \seqsplit{Interactive\_Narrative\_Engine} \\
Q11 & ``I need an AI capable of conducting tarot card readings---interpreting card combinations based on spread positions and delivering symbolically rich, layered interpretations.'' & \seqsplit{Occult\_\&\_Ritual\_Systems} & --- \\
Q12 & ``Help me build a multi-agent AI workflow where each agent handles a different stage of the pipeline, and context must be reliably passed between agents without loss.'' & \seqsplit{Distributed\_Cognition\_\&\_Context\_Orch.} & \seqsplit{Autonomous\_System\_Engineering} \\
Q13 & ``I need a thinking framework for analyzing competitor weaknesses and identifying systemic vulnerabilities to inform business strategy.'' & \seqsplit{Adversarial\_Systems\_Thinking} & \seqsplit{Strategic\_Decision\_Frameworks} \\
Q14 & ``Design an AI that can automatically convert experimental code results into methodology and results sections formatted for academic publication.'' & \seqsplit{Academic\_Research\_Synthesis\_Pipeline} & \seqsplit{Academic\_Insight\_\&\_Forensics} \\
Q15 & ``I want to build a sales conversation system that can identify a prospect's purchase intent and guide the conversation toward closing in the most persuasive way possible.'' & \seqsplit{Revenue\_Generation\_\&\_Commercial\_Logic} & \seqsplit{Product\_Psychology} \\
Q16 & ``Help me design an AI assistant that can detect a user's cultural background and automatically adjust its communication style and content framing accordingly.'' & \seqsplit{Cross\_Cultural\_Localization\_Intelligence} & \seqsplit{Persona\_\&\_Narrative\_Synthesis} \\
Q17 & ``I need an AI system that can continuously optimize its own prompt logic---learning from past failures and iteratively improving its internal reasoning.'' & \seqsplit{Self\_Evolution\_\&\_Refinement} & \seqsplit{Meta\_Data\_\&\_Engineering} \\
Q18 & ``Design an interactive fiction engine where every player choice shapes the world's development---with multiple branching storylines that remain internally consistent.'' & \seqsplit{Interactive\_Narrative\_\&\_Fiction\_Engine} & \seqsplit{RPG\_Narrative\_Director} \\
Q19 & ``I want to validate a business idea from scratch at minimum cost---testing market response before committing to any development.'' & \seqsplit{Minimalist\_Entrepreneurship\_Execution} & \seqsplit{Community\_Led\_Business\_Inception} \\
Q20 & ``Help me design a sound design AI that, given a visual scene description, can derive the corresponding sound layers and acoustic landscape structure.'' & \seqsplit{Sensory\_Audio\_Design} & --- \\
\end{longtable}

\begin{table}[h]
\centering
\caption{Representative distractor categories by interference tier.}
\label{tab:distractor_sample}
\small
\begin{tabular}{p{0.04\linewidth} p{0.35\linewidth} p{0.50\linewidth}}
\toprule
\textbf{Tier} & \textbf{Distractor Name} & \textbf{Target Category Confused With} \\
\midrule
High & \seqsplit{Agent\_Handoff\_Protocol\_Design} & \seqsplit{Distributed\_Cognition\_\&\_Context\_Orchestration} \\
High & \seqsplit{Competitive\_Intelligence\_Synthesis} & \seqsplit{Adversarial\_Systems\_Thinking} \\
High & \seqsplit{Conversion\_Funnel\_Architecture} & \seqsplit{Revenue\_Generation\_\&\_Commercial\_Logic} \\
High & \seqsplit{Code\_To\_Methodology\_Translator} & \seqsplit{Academic\_Research\_Synthesis\_Pipeline} \\
High & \seqsplit{Cultural\_Signal\_Detection} & \seqsplit{Cross\_Cultural\_Localization\_Intelligence} \\
High & \seqsplit{Recursive\_Self\_Audit\_Engine} & \seqsplit{Axiomatic\_Logic\_\&\_Audit\_Systems} \\
High & \seqsplit{Dynamic\_Difficulty\_Calibration} & \seqsplit{Interactive\_Pedagogy\_\&\_Gamification} \\
\midrule
Low & \seqsplit{Mycological\_Network\_Design} & (none---pure volumetric noise) \\
Low & \seqsplit{Trophic\_Cascade\_Analyzer} & (none---pure volumetric noise) \\
Low & \seqsplit{Glacial\_Erosion\_Patience\_Model} & (none---pure volumetric noise) \\
Low & \seqsplit{Coral\_Reef\_Ecosystem\_Modeler} & (none---pure volumetric noise) \\
Low & \seqsplit{Bioluminescence\_Signal\_Design} & (none---pure volumetric noise) \\
\bottomrule
\end{tabular}
\end{table}

\section{Experimental Prompt Templates}
\label{app:prompts}

\paragraph{Version A and B System Prompt (minimal framing):}

\begin{verbatim}
You are a professional Prompt Engineer.
I will provide you with a skills.json library and 20 task requirements.
For each task, select the most relevant skills from the library.
For each skill you select, list ONLY the category_name and skill_name.
Do NOT explain your reasoning yet — just list the selections for all
20 tasks first.
Ensure that every skill you reference actually exists in the library.
\end{verbatim}

\paragraph{Version C Additional Prompt Content (added to above):}

\begin{verbatim}
Before reading the library:
- Scan the category_description fields to understand each category's scope.
- Priority rule: when a broad pipeline/orchestration/governance category
  and a narrow mechanism/component category both seem relevant,
  prefer the broader one.
- Key high-priority categories: Cognitive_Architecture_&_Routing (routing),
  Axiomatic_Logic_&_Audit_Systems (logic/audit, NOT narrow audit loops),
  Distributed_Cognition_&_Context_Orchestration (multi-agent governance),
  Adversarial_Systems_Thinking (thinking frameworks, NOT intelligence),
  Academic_Research_Synthesis_Pipeline (end-to-end pipeline),
  Revenue_Generation_&_Commercial_Logic (commercial system),
  Cross_Cultural_Localization_Intelligence (full localization).
\end{verbatim}

\paragraph{Version D:} Uses Version~B's JSON file (with \texttt{\_summary} block) together with Version~C's extended system prompt, producing the dual-layer configuration.

\section{Distractor Category Design}
\label{app:distractors}

Table~\ref{tab:distractor_sample} shows representative distractor categories

\end{document}